\documentclass{article}
\usepackage{spconf,amsmath,graphicx}
\usepackage{xcolor}
\usepackage{array}
\usepackage{url}
\graphicspath{ {./images/} }

\title{Building Height Estimation Using Shadow Length in Satellite Imagery}
\name{Shayaan Chaudhry$^*$, Mahd Qureshi$^*$, Sana Jabbar, Murtaza Taj}
\address{Computer Vision and Graphics Lab, LUMS}
\begin{document}
%\ninept
%
\maketitle
\def\thefootnote{*}\footnotetext{These authors contributed equally to this work}
\begin{abstract}
Estimating building height from satellite imagery poses significant challenges, especially when monocular images are employed, resulting in a loss of essential $3$D information during imaging. This loss of spatial depth further complicates the height estimation process. We addressed this issue by using shadow length as an additional cue to compensate for the loss of building height estimation using single-view imagery. We proposed a novel method that first localized a building and its shadow in the given satellite image. After localization, the shadow length is estimated using a regression model.
To estimate the final height of each building, we utilize the principles of photogrammetry, specifically considering the relationship between the solar elevation angle, the vertical edge length of the building, and the length of the building's shadow.
For the localization of buildings in our model, we utilized a modified YOLOv7 detector, and to regress the shadow length for each building we utilized the ResNet18 as backbone architecture.
Finally, we estimated the associated building height using solar elevation with shadow length through analytical formulation. We evaluated our method on $42$ different cities and the results showed that the proposed framework surpasses the state-of-the-art methods with a suitable margin. 
Our code is available at: https://github.com/nullptr-code/building-height-model.git. 

\end{abstract}
\section{Introduction}
\label{sec:intro}
Accelerating the pace of urbanization results in urban sprawl and creates immense pressure on urban land use and resources.
More than $68\%$ of the world population will live in urban areas by the year $2050$ \cite{worldpopulation}.
Integrated planning and management regarding the development of urban areas is one of the biggest challenges in the world today.
Vertical expansion of urban areas can be a solution to shelter the growing urban population \cite{vertical-expansion}.
But this results in high-rise buildings that have already been established in big cities.
Building height estimation is essential urban geographic information for the assessment of land changes \cite{ijgi2022}, smart cities modeling \cite{Perera2021}, and autonomous driving \cite{Sautier2022}.
The estimation of building heights is always been challenging in the field of remote sensing \cite{building_height_with_shadow}.
Building heights can be estimated using light detection and ranging (LiDAR) \cite{Retrieving-building-height-LiDAR,Extract-Building-LiDAR-Data} and synthetic aperture radar (SAR) \cite{Height-Estimation-for-SAR,height-retrieval-from-single-SAR}.
%--------------------
\begin{figure}[t]
	\centering
{\includegraphics[width=1.1\columnwidth]{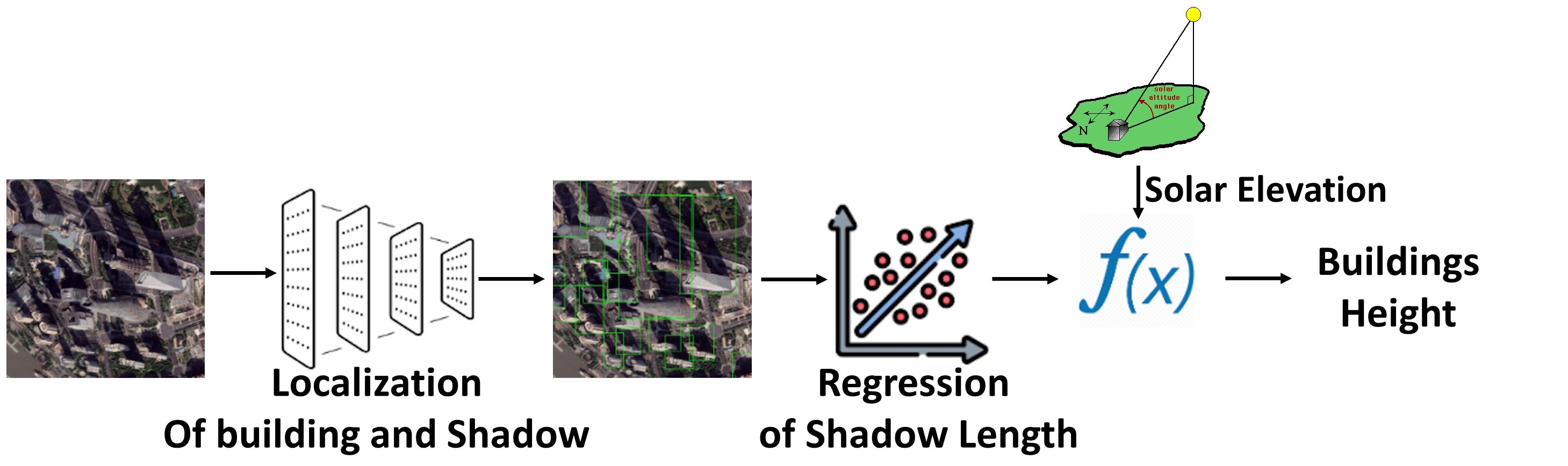}}
	\caption{Overview of the proposed framework for building height estimation using shadow length.}
	\label{fig:overview}
 \vspace{-0.5cm}
\end{figure}
%\vspace{-0.75cm}
%-----------------------
\begin{figure*}[t]
	\centering
{\includegraphics[width=0.8\textwidth]{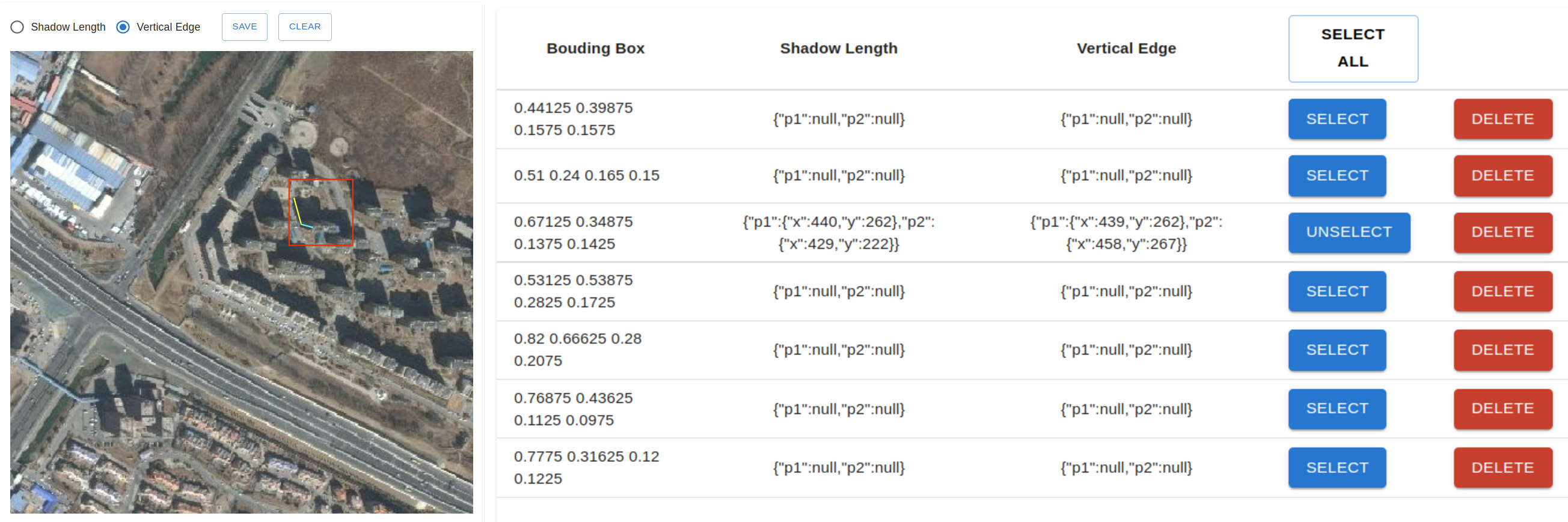}}
\caption{Sample image showing the screenshot of annotation tools. Bounding boxes along with the annotations for vertical edge and shadow length for the encompassing building are shown for each bounding box.}
	\label{fig:annotationtool}
\end{figure*}

Soergel et al. proposed a segmentation-based building height estimation method using SAR imagery.
Wegner et al. proposed building height estimation using a pair of SAR and single optical imagery \cite{wegner2013combining}.
Brunner et al. proposed SAR based building height estimation method and evaluated over flat and  gable building roofs  \cite{brunner2009building}. 
These methods performed building height estimation with high efficiency while utilizing multi-sensor data which is constrained by high computational resources and  acquisition cost \cite{Min-lu2022}.
The above-mentioned difficulty can be handled using high-resolution optical satellite imagery.
Many building heights can be estimated with optical satellite imagery using deep learning methods \cite{Yinxia2021, karatsiolis}.
But deep learning models perform end-to-end mapping from optical satellite image to height map.
Here our goal is to utilize the mathematical models based on shadow length along with deep learning models.
Unlike, data-driven end-to-end based height estimation~\cite{RouxISPRS2022}, there are three major formulations to compute the height of an object using imagery~\cite{Hartley:2003:MVG:861369}. This includes i) radial displacement method, ii) parallax method, and iii) shadow method.
Qi et al. \cite{Qi2016} utilized shadow length and vertical edge length-based celestial geometry formulation using Google Earth Imagery.
Qi et al. \cite{Qi2016} performed an evaluation over a local scale for $21$ buildings in Lin’an.
Whereas, we performed experimentation over densely populated cities in China.
Secondly, we combined the detection model and regression networks to get the final building heights.
We designed an annotation tool to mark the shadow lengths of buildings in images and passed that shadow length to a mathematical model to get the final building height.
Our key contributions to this study are listed as follows:
\begin{itemize}
    \item We developed a localization and regression-based novel frame for building height estimation using monocular satellite imagery.
    \item We developed a data annotation tool that annotates buildings and shadow lengths. 
    We have developed a new dataset for the remote  sensing community.
    \item We also incorporated mathematical formulation along with our regression model to get improved building height estimation results.
    \item When we evaluated our proposed framework on densely populated $42$ cities of \uppercase{C}hina, our model surpassed the performance of state-of-the-art methods.
\end{itemize}
The remainder is organized as Section 2 regarding the dataset, Section 3 describes the detailed methodology, results are reported in Section 4, and in the end, the conclusion is drawn in Section 5.

\section{Dataset}
Most of the datasets developed so far for building height estimation consist of height maps only and are suitable for end-to-end methods only~\cite{Yinxia2021}. Thus, we prepared our own dataset that contains a building height map, bounding boxes, latitude and longitude, and shadow length for each building. We prepared our dataset by extending the popular $42$ Chinese cities dataset~\cite{Yinxia2021}. This dataset provides pixel-wise labeling for geo-localized images patched each having resolution of $400 \times 400$ pixels, where each image corresponds to a $1000 \times 1000 $ m region. This dataset provides us with the number of floors and not the true height labels. To cater to this, we multiply the number of floors by $3$m to convert it to a height map. We randomly selected over $3000$ images from the datasets some of which were later filtered as being inappropriate for our model. Most of the images that were flagged were because there were no buildings visible in them resulting in total $2704$ images. In total, we were able to annotate shadow lengths and bounding boxes for approximately 50,000 buildings.

\subsection{Annotation Tool}
We used a popular tool known as DarkLabel~\cite{darklabel} for marking the bounding boxes of each building in an image. To mark the shadow lengths, we developed our own annotation tool. Our tool takes as input a list of bounding boxes associated with images in YOLO format~\cite{wang2022yolov7}. It allows the user to mark the start and end point of the shadow for each building/bounding box and stores them in a database. Furthermore, for each annotated building, we store the latitude, longitude, and the time when the image was taken for the calculation of solar elevation angle which is required for height calculation in the mathematical formula. 

Furthermore, we used the analytical method discussed in section\ref{sec:height_estimate_sl} to compute the estimated height of the building and compare it with the ground truth to observe the errors made by the annotators during the annotations. Only date information is available from the freely available images in Google Earth which is insufficient for the calculation of solar elevation angle. We solved this issue, during the annotation process, by calculating the estimated time of imaging through a minimization process. Our minimization selects the times of the date that provides the least error in building height as compared to ground truth. A screenshot of the developed annotation tool and annotation process is shown in Fig.~\ref{fig:annotationtool}.

\subsection{Dataset Analysis}
\begin{figure}[t]
    \centering
         \begin{tabular}{cc}
            \includegraphics[width=0.45\columnwidth]{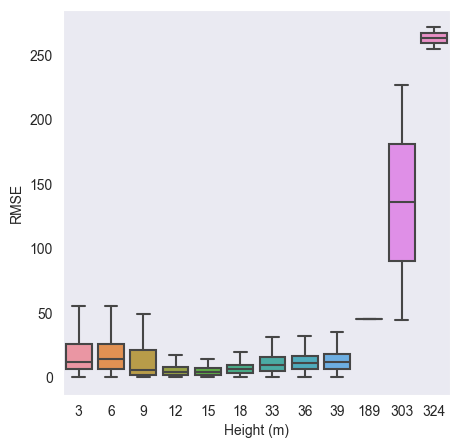} &
            \includegraphics[width=0.45\columnwidth]{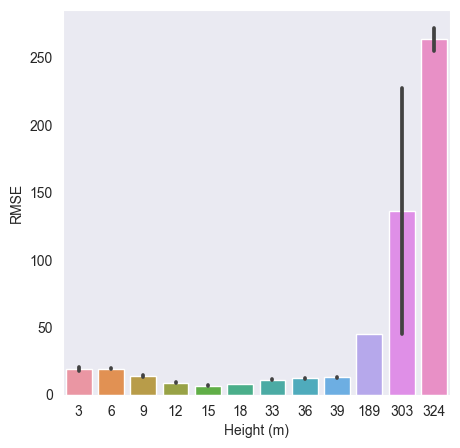}\\
            (a) & (b)\\
         \end{tabular}
    \caption{(a) Box plots of Root Mean Square Error on the dataset, plotted across values of ground truth height. (b) Bar plot representing the average (mean) Root Mean Square Error plotted against values of ground truth height. We can observe that the range of values that RMSE takes is small for buildings lying in the $12$-$30$m range. The range of RMSE for buildings in the height range of $3$-$9$m is pretty large which suggests noise. Moreover, buildings with a height $>30$m show very large RMSE.}
    \label{fig:dataset_graphs}
\end{figure}

Since our dataset derives from previously available dataset~\cite{Yinxia2021}, there was noise due to the quantization of height into building floors and imbalance in height labels. Using shadow lengths we marked using our custom annotation tool, we attempted to calculate the height of buildings in the dataset to see how viable our method would be before implementing the DNN.  We identified the causes of these noises and catered to them in the ways outlined in this section. 
Figure~\ref{fig:dataset_graphs}  shows the average errors associated with the ground truth and predicted building heights. We can see that buildings with heights greater than $30$, and significantly greater errors than buildings with heights less than or equal to $30$. To avoid this we added a cap of $30$m to heights. We used the label '$33$' to represent buildings with heights greater or equal to $30$. 

Moreover, we noticed that for buildings with heights labeled $9$ or lower, there were instances where very large shadow lengths were marked. This was obviously noisy data because it does not make sense for our buildings of height $3$m to $9$m to project very large shadows. For this reason, we excluded instances where shadow length was marked greater than or equal to $50$ for these buildings. 
These adjustments brought the mean error significantly lower across the dataset. However, there was still significant noise that we could not account for. Overall, the Average error across the dataset was approximately 10m or ($3$-$4$ floors). 

%We only had the date but not the time from Google Earth. Calculated our predictions for all the images and selected the time which would give the least height error. 

%. Our annotation tool takes in the bounding boxes annotations in YOLO format and shows them on a canvas to the annotator. The annotator is able to select each bounding box, mark the start and the end of the shadow length, and save the labels in our MongoDB database.

%\textbf{Points yet to add} \\
%Overlayed bounding boxes on masks from original dataset to get ground truths for height values. \\
%We only had the date but not the time from Google Earth. Calculated our predictions for all the images and the selected the time which would give the least height error. \\

%

%================================================
\section{Methodology\label{sec:methodology}}
%===============================================
Our height estimation framework is designed around three integral components: (i) building localization, (ii) shadow length estimation, and (iii) height estimation, which incorporates a novel shadow-based height loss. Initially, we utilize YOLOv7 to train an object detector that precisely delineates a bounding box around each building. Following this, each localized building undergoes a detailed analysis to determine the length of its shadow, which is crucial for the subsequent estimation of the building's height. Our deep neural network (DNN), which employs ResNet18 as its core architecture, is tasked with regressing the shadow length for each building. Finally, we apply our innovative shadow-based height loss technique to accurately derive both the shadow length and the corresponding building heights.
%==============================
\subsection{Building and Shadow 
Localization\label{subsec:buildingLocalization}}
%==============================
In order to identify buildings along their shadows we employed modified YOLOv7 model~\cite{wang2022yolov7}. This model efficiently identifies each building and its shadow within a bounding box generated during the detection phase. Following the initial detection, the bounding boxes produced by YOLOv7 are processed through our specialized Cropping model. This subsequent model takes the initial image, crops it according to the dimensions specified by the bounding boxes, and resizes the cropped images to a standard resolution of 
$(50, 50, 3)$. These resized images are then formatted appropriately for input into our Regression Model, facilitating further analysis. The following section~\ref{subsec:height_estimate_sl} describes how height is being compressed using these cropped images of buildings.
%=====================
\subsection{Height Estimation via Shadow Length}
\label{subsec:height_estimate_sl}
%=====================

Upon successfully employing YOLOv7 for the initial detection of buildings and their accompanying shadows, each identified image is meticulously cropped to isolate the specific area containing both the building and its shadow. This focused segmentation is essential for precise analysis in the subsequent stages of our methodology. The cropped images are then input into a regression model specifically trained to estimate the length of the shadows. This regression process is crucial as it quantitatively assesses the shadow dimensions, which are pivotal for accurate height estimation of the buildings based on shadow analysis. The precise measurement of shadow length by our regressor enables a robust foundation for deriving building heights through advanced geometric and photometric techniques.
Height estimation via shadow requires the use of only a single image whereas the other two approaches require the availability of stereo pair. To calculate the height of a building, we make use of the length of a shadow that the building produces. This can be achieved with the following formula:
where $H$ is the height of a building, $S_l$ is the corresponding length of shadow and $\sigma$ is the solar elevation angle~\cite{REDA2004577}. In this work, we use shadow length as a proxy for building height and propose a novel deep-learning architecture that combines the close-from analytical approach within a learning paradigm. This results in a more meaningful and interpretable estimation of height backed up with the mathematical model. The proposed model is discussed next.
%----------------------
\begin{figure*}[ht]
    \centering
         \begin{tabular}{cc}           \includegraphics[width=0.95\columnwidth, trim={0 23cm 0 0},clip]{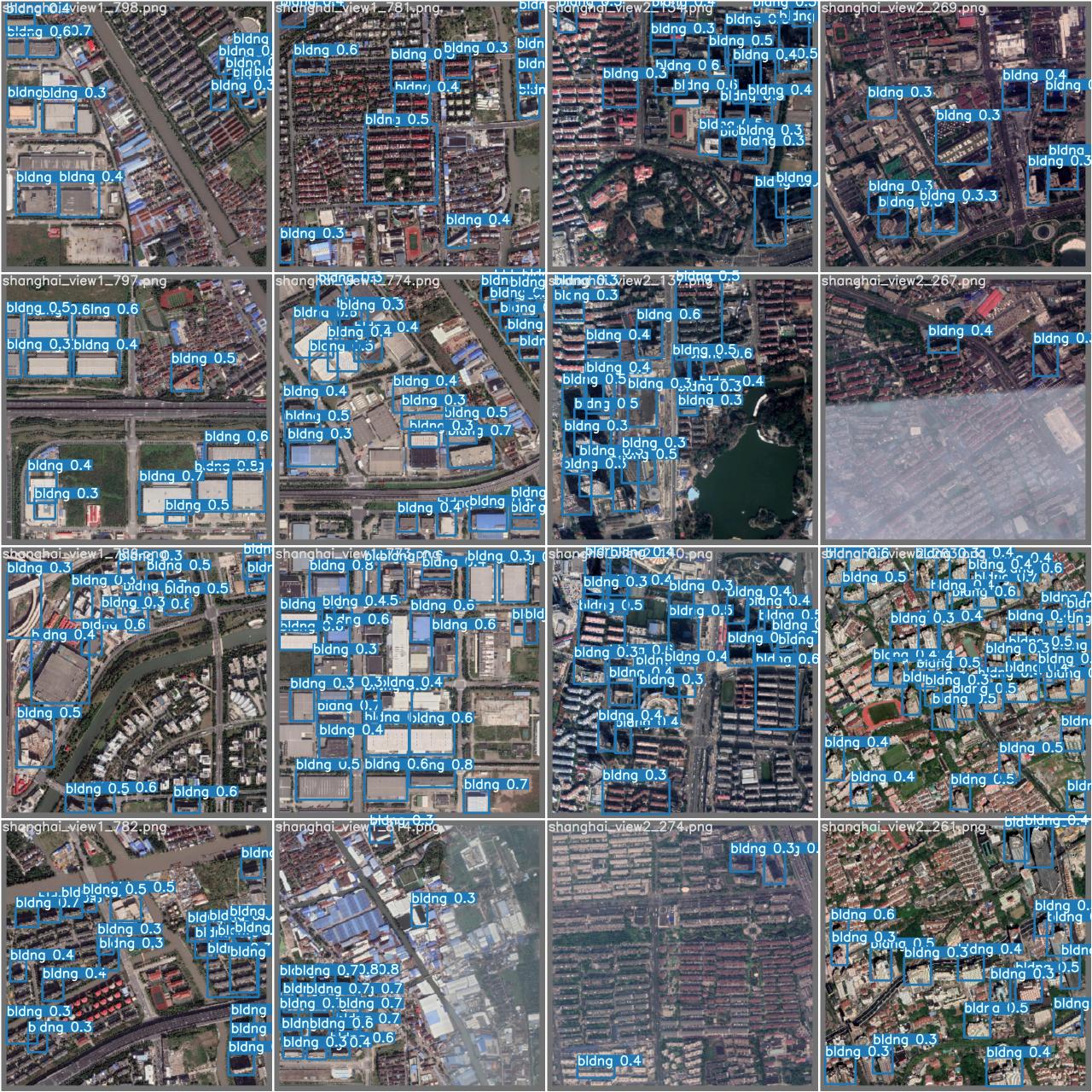} &
          \includegraphics[width=0.95\columnwidth, trim={0 23cm 0 0},clip]{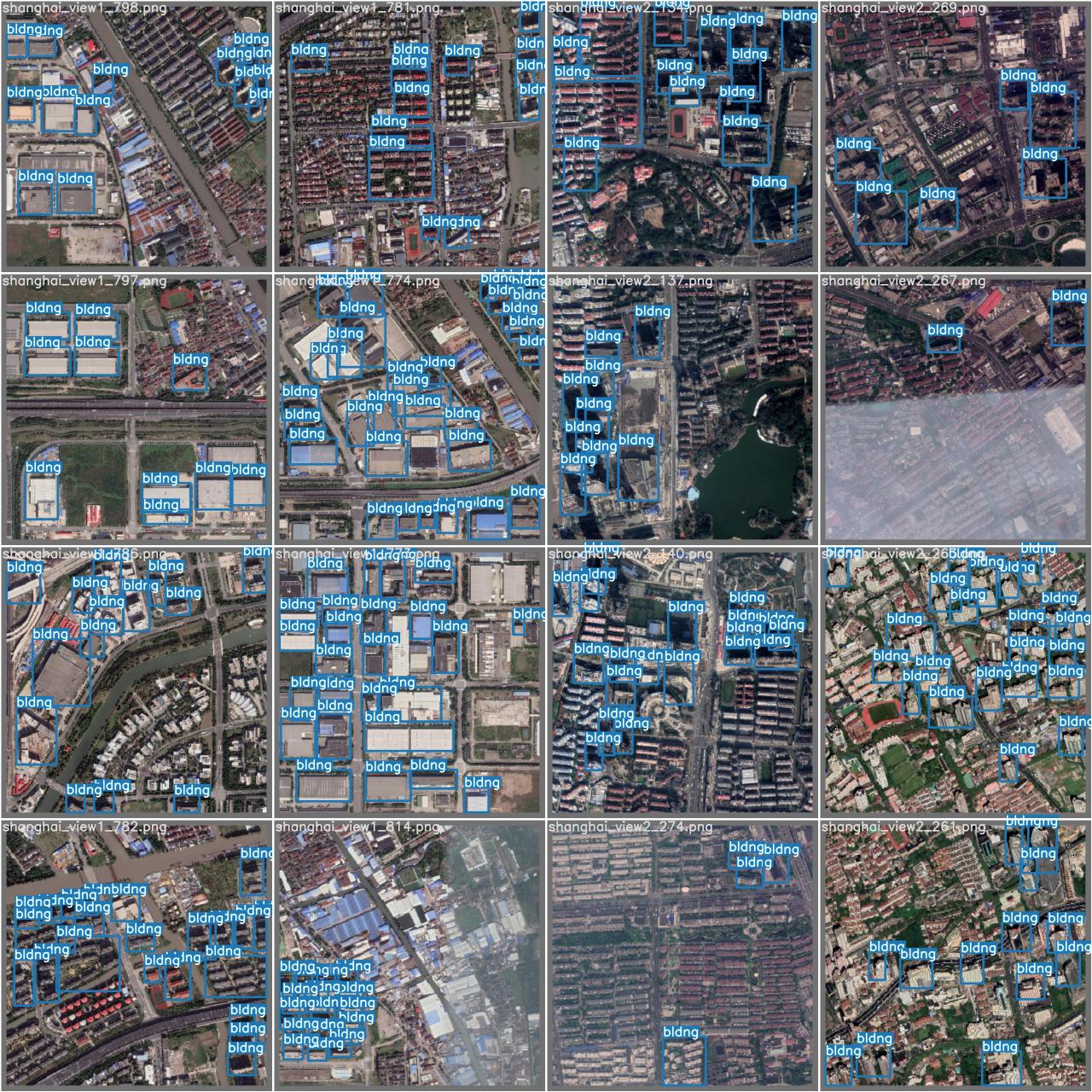}\\
            (a) & (b) 
         \end{tabular}
    \caption{(a) YOLOv7 Predictions (b) Bounding box ground truth. }
    \label{fig:yolov7}
\end{figure*}
%-----------------------
%
\begin{equation}
    H = {S_l}{ \tan (\sigma)}.     
\end{equation}
\begin{table}[b]
\begin{center}
    \caption{Table representing experiments with hyperparameters showing RMSE (in meters) in height and shadow length.}
    \begin{tabular}{ m{0.8cm} | m{1.2cm}| m{0.9cm} | m{1cm} | m{1.2cm} | m{1.2cm}}
         \hline
         Loss & Optimizer & $\eta$ & Weight Decay & Height Loss (m) & Shadow Loss (m)\\ 
         \hline
         \hline
         L1 & Adam & 0.0001 & $1\times e^{-5}$ & 3.84 & 4.4\\  
         L1 & SGD & 0.0001 & $1\times e^{-5}$ & 11.1 & 14.5\\  
         MSE & Adam & 0.0001 & $1\times e^{-5}$ & 4.63 & 5.46 \\  
         MSE & SGD & 0.0001 & $1\times e^{-5}$ & 16.7 & 80.2 \\  
         \hline
    \end{tabular}
\end{center}
\label{tab:1}
\end{table}

%--------------
\section{Results and Evaluation}
We evaluated our proposed method on $42$ cities dataset~\cite{Yinxia2021} that we extended by adding required annotations for building location and shadow length. The method was compared with best performing algorithm on this dataset i.e. MM$^3$Net~\cite{Yinxia2021} as well as with baseline random forest classifier.

\subsection{Implementation Details}
In order to find appropriate hyperparameters, we ran multiple experiments as shown in Table~1. These results are on a test dataset of $500$ buildings. It can be seen that results using L1 loss with Adam optimizer and having learning rate $\eta$ of $0.0001$ with weight decay $1\times e^{-5}$ results in the smallest value of Root Mean Square Error (RMSE). We used these hyperparameters for the rest of the experiments.
For testing, we randomly sampled from the full dataset, a smaller dataset of around $5000$ instances that contained only buildings with heights less than or equal to $30$m. We extracted only those buildings where we obtained RMSE of $2.5$m or less from our mathematical formula (calculated from the ground truth label of shadow length).
%------------------------------------
\subsection{Comparative Analysis}
%-----------------------------------
We compared our method with the best-performing method on the $42$ cities dataset i.e. with MM$^3$Net. We also compared our method with the baseline random forest classifier. The comparative analysis in terms of root mean square error is shown in Table~\ref{tab:2}. It can be seen that our proposed method outperformed both the MM$^3$Net and random forest. It should also be noted that MM$^3$Net used both optical as well as multi-spectral imagery whereas in our approach we only use optical imagery. Despite this, our method achieved a minimum RMSE of only $3.84$ as compared to $6.259$ achieved by MM$^3$Net. In other words, our method reduced the error in building height by $2.419$ meters.
%---------------------
\begin{table}[t]
\begin{center}
    \caption{Table showing comparative analysis with state-of-the-art methods for $42$ cities. Root mean square error is reported in meters (m).}
    \begin{tabular}{m{8em}| m{4cm}|m{1cm} } 
          \hline
          Model &  Imagery & RMSE (m)\\ 
          \hline
          RF~\cite{Yinxia2021} & Multi-view, Multi-spectral & $7.46$\\ 
          MM$^3$Net~\cite{Yinxia2021} & Multi-view, Multi-spectral & $6.26$ \\ 
        Stereoential Net~\cite{jabbar2023stereoential} &
            Multi-view&
		$ 5.95$\\
        Stereollax Net~\cite{jabbar2024stereollax} &
            Multi-view&
		$ 5.74$\\
          Our Model &  Monocular &\bf{$3.84$} \\          
          \hline
        \end{tabular}
\end{center}
\label{tab:2}
\end{table}
%-------------------------
Figure~\ref{fig:yolov7} shows the qualitative evaluation of the YOLOv7 detector that we trained on 42 cities remote sensing dataset. This figure also provides a comparative analysis with ground truth. It can be seen that our detections highly overlap with annotation ground truth, thus we not only localize each building we also estimate its height with the lowest error in height.

%For testing, we randomly sampled from the full dataset, a smaller dataset of around $5000$ instances which contained only contained buildings with height less than or equal to $30$m. We further extracted only those buildings where we obtained Root Mean Square Error of $2.5$m or less from our mathematical formula (calculated from ground truth label of shadow length). 

%Some results from various parameters on this test dataset are shown in the Table~\ref{tab:1}. It can be seen that results using L1 loss with Adam optimizer results in smallest value of root mean squared error.

%Results without deep unfolding/end-to-end
%Results with deep unfolding/with maths

%\subsection{Quantitative Evaluation}
%\textcolor{red}{Comparison with several methods that Sana is 
%uses Table showing comparison}

%\subsection{Qualitative Evaluation}

% -------------------------------------------------------------------------
%
\section{Conclusions}
In this work, we addressed the problem of building height estimation by leveraging both learning-based techniques and mathematical principles. Firstly, we implemented a localization method to identify the building instances based on their shadows. Then, we utilized regression analysis to estimate the length of the shadows, which played a pivotal role in determining the final building height estimation. 
This allowed training the model based on data as well as via mathematical modeling resulting in a more intuitive solution. 
This integrated approach enabled us to achieve accurate and reliable results in building height estimation.
\begin{figure}[t]
	\centering
{\includegraphics[width = 0.4\columnwidth]{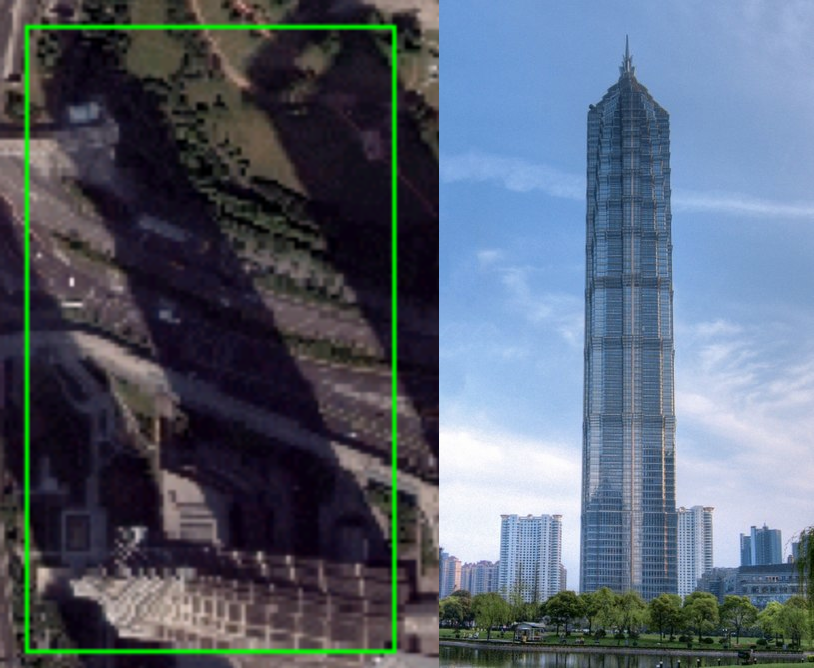}}
	\caption{Jin Mao Tower in Shanghai. Actual height $420$ meters. The height computed from our annotated shadow length is $430$ meters.}
	\label{fig:tower}
\end{figure}
We evaluated this hybrid approach on a $42$-cities dataset by adding bounding boxes, shadow length, and building vertical edge length annotations. Our proposed model back propagates error first through the analytical model and then through the DDN layers resulting in hybrid learning. In the future, we aim to extend our method to the radial displacement method and parallax-based formulations, as well as extend the evaluation to other methods and datasets.

\bibliographystyle{IEEEbib}
\bibliography{strings,refs}
\newpage

\end{document}